\documentclass{article}
\usepackage[utf8]{inputenc}
\usepackage{xcolor}
\usepackage{amsmath}
\usepackage{graphicx}
\usepackage{float}
\usepackage{amsfonts}
\usepackage{caption}
\usepackage{array}
\usepackage{subcaption}
\usepackage{multirow}
\usepackage{natbib}
\usepackage{color}
\usepackage{hyperref}
\usepackage[left=3cm,right=3cm,top=3cm,bottom=3cm]{geometry}

\DeclareMathOperator*{\argmin}{argmin}

\title{Frugal day-ahead forecasting of multiple local electricity loads by aggregating adaptive models}
\author{Guillaume Lambert\thanks{Électricité de France R\&D, Palaiseau, France, \href{mailto:lambert\_guillaume@outlook.fr}{lambert\_guillaume@outlook.fr}} , Bachir Hamrouche\thanks{Électricité de France R\&D, Palaiseau, France, \href{mailto:bachir.hamrouche@edf.fr}{bachir.hamrouche@edf.fr}} \ and Joseph de Vilmarest\thanks{Viking Conseil, Paris, France, \href{mailto:joseph.de-vilmarest@vikingconseil.fr}{joseph.de-vilmarest@vikingconseil.fr}}}

\date{\today}

\begin{document}

\maketitle

\begin{abstract}
We focus on day-ahead electricity load forecasting of substations of the distribution network in France; therefore, our problem lies between the instability of a single consumption and the stability of a countrywide total demand. Moreover, we are interested in forecasting the loads of over one thousand substations; consequently, we are in the context of forecasting multiple time series. To that end, we rely on an adaptive methodology that provided excellent results at a national scale; the idea is to combine generalized additive models with state-space representations. 
However, the extension of this methodology to the prediction of over a thousand time series raises a computational issue. We solve it by developing a frugal variant, reducing the number of parameters estimated; we estimate the forecasting models only for a few time series and achieve transfer learning by relying on aggregation of experts. It yields a reduction of computational needs and their associated emissions.
We build several variants, corresponding to different levels of parameter transfer, and we look for the best trade-off between accuracy and frugality. The selected method achieves competitive results compared to state-of-the-art individual models. Finally, we highlight the interpretability of the models, which is important for operational applications.
\end{abstract}

\section{Introduction}\label{section:introduction}

Electricity consumption forecasting is essential for numerous activities: managing the electricity network, investment and production planning, trading on the electricity markets, and reducing power wastage. To suit these activities, forecasts are performed at different horizons, from short-term (hours, days) to long-term (years), and on different scales, from individual to national. New variability in the electricity load has recently emerged due to several factors: new decentralized production units (like solar panels), new actors with the opening of the French electricity market, new uses (like electric mobility), and the COVID-19 pandemic. They bring essential changes in the electricity consumption, and we argue that forecasting models must be updated to take them into account.

Numerous forecasting approaches have been proposed in the past before this new variability, and we refer to the Global Energy Forecasting Competitions (GEFCom) for an overview \citep{hong2014global,hong2016probabilistic,hong2019global}. Classical time series methods such as auto-regressive integrated moving-average (ARIMA, \cite{huang2003short, chodakowska2021arima}) and exponential smoothing \citep{abd2013electricity} have been used to forecast at the very-short term (hours-ahead). Machine learning approaches provide better forecasts using explanatory variables, especially for larger horizons. Indeed, judging that electricity demand is a human activity, it can be predicted using explanatory variables, the most important ones being weather and calendar data.
In particular, Generalized Additive Models (GAMs, \cite{wood2017generalized}) have been widely applied to forecast the electricity consumption \citep{pierrot2011short,goude2013local,fasiolo2021fast}.

More recently, a state-space approach has been investigated to adapt GAMs over time \citep{de2022modeles}; this online model allows GAMs to adapt to new variabilities, such as the COVID-19 pandemic, to improve forecasts \citep{obst2021adaptive,de2022state}. Finally, the variety of models motivates predicting the demand with a combination of forecasters, yielding a final prediction better than any individual model; that is the aggregation of experts \citep{cesa2006prediction}, which has also shown good results on load forecasting \citep{gaillard2015forecasting}. Aggregation of experts is an online method which combines predictions given by experts in a weighed sum where the weights evolve over time.

In this paper, we study day-ahead electricity load forecasting on the distribution network in France. The electricity consumption is measured on about 2,200 substations located at the frontier between the high voltage grid and the distribution network. More precisely, we have access to the data of 1,344 of them. We thus consider forecasting at a local scale.
The high number of substations motivates the need for a methodology to forecast a large number of time series efficiently; we look for the right trade-off between good accuracy and frugality in the number of parameters and computational complexity.
Frugal machine learning can be motivated by economic reasons, for instance to avoid costly data collection or transmission, and by environmental goals like the reduction of computers energy consumption. An example where frugality is key is the implementation of machine learning algorithms in portable devices such as smartwatches \citep{evchenko2021frugal}: transmission of data is difficult and intensive use of machine learning algorithms is limited by computational and battery saving constraints. In our case, the limitation comes from the learning process which becomes infeasible in a reasonable time with such a large number of time series to predict.

\begin{figure}
     \centering
     \includegraphics[width=0.8\textwidth]{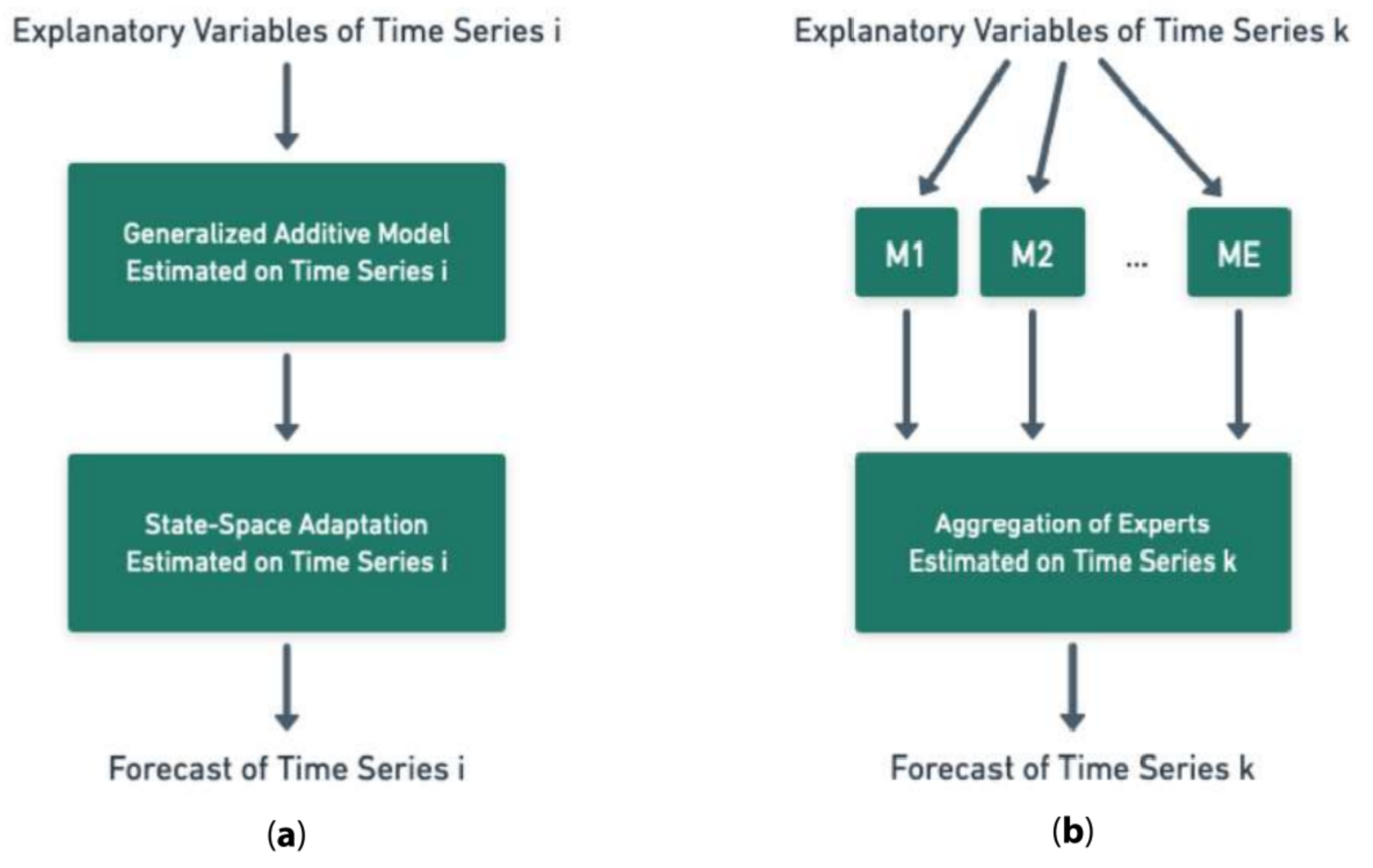}
        \caption{Transfer learning through aggregation of experts in two steps : (\textbf{a}) we train individual models for $E$ time series for a small $E$ compared to the number of time series, then (\textbf{b}) to forecast any time series $k$, we apply models $M_1$, $M_2$, \dots, $M_E$ and we combine all the resulting forecasts using aggregation of experts.}
        \label{fig:diagrams}
\end{figure}

We present this challenge from a transfer learning point of view that we represent with diagrams in Fig. \ref{fig:diagrams}.
We train individual models on a few time series, relying on GAMs and their adaptive variant based on state-space models. Then, we transfer these models learned on a small fraction of the time series to all of them, instead of training an individual model for each time series. We use aggregation of experts as a transfer tool. Precisely, we apply our base models on each time series even if the models were not trained on them, and then aggregating these weak forecasters yields a procedure able to scale to a large number of time series. We build several aggregations involving different kinds of models, where transfer learning occurs at multiple levels.
We discuss the computational cost for each aggregation type and focus on their frugal nature. Indeed, optimization of individual models to a large number of time series is very costly, especially the performing models we want to use, while aggregation of experts is not. We show that our methods compete with individual models that we detail, even during the first French lockdown due to COVID-19, while achieving low computational cost. Transfer learning through aggregation of experts has also been studied in a hierarchical setting, using regional load data as sources and national load data as target \citep{gaucher2021hierarchical}.

Following this introduction, Section \ref{section:methodo} establishes the methods used in our work. In Section \ref{section:dataModelPresentation} we introduce the data and the models applied. We analyze and present in Section \ref{section:experiments} the results of the forecasting tasks. Finally, we conclude the paper and propose future work.

\section{Methodology}\label{section:methodo}

In this section we first characterize our problem in the transfer learning framework. The transfer is obtained by aggregation of experts, and we describe its principle. Finally, we introduce the different models we consider and the resulting aggregation methods.

\subsection{Transfer learning context}\label{subsection:TL context}

We start by expressing the transfer learning context using the definitions and vocabulary used for transfer learning methods \citep{zhuang2020comprehensive}. First, we define the notion of domain and task. A domain $\mathcal{D} = \{ \mathcal{X}, P(X) \}$ is composed of a feature space $\mathcal{X}$ and a marginal distribution $P(X)$ of an instance set $X = \{ x \mid x_i \in \mathcal{X}, i = 1, \dots, D \}$. A task $\mathcal{T} = \{ \mathcal{Y}, f \}$ is composed of a label space $\mathcal{Y}$ and a decision function $f$. Given $m_S \in \mathbb{N}^+$ source domains and tasks, and target domain and task, transfer learning is the use of the knowledge within the source domains and tasks to improve the performance of the target decision function.

Each substation represents one target whose decision function varies from the other substations' decision functions. Thus, we have $m_T = 1,344$ transfer learning tasks, each corresponding to one forecasting task. It results in a multi-task transfer learning situation. On the other hand, judging that we will use several sources to do the transfer, we have $m_S > 1$, resulting in a multi-source transfer learning method.

In our case, we are in a homogeneous and inductive transfer learning situation. Indeed, a homogeneous transfer learning scenario occurs when the source and target feature spaces are equal, as well as the source and target label spaces. The explanatory variables (calendar and meteorological variables and past electricity load) are equally available for all substations and range thus in the same feature space. The mutual label space is $\mathbb{R}_+$ corresponding to the electricity measurement range. Concerning inductive transfer learning, it happens when labeled data are available in the target domain to induce the target decision functions, which is our case.

\subsection{Transfer learning using aggregation of experts}\label{subsection:aggregationPrinciple}

A natural idea would be to determine a subsample of $m_S$ data sets, build one individual model per data set, and select the best among the $m_S$ models for each substation we want to forecast. The information on the best model is unavailable; one way to estimate it is through the aggregation of experts.

In the aggregation of experts context \citep{cesa2006prediction}, we want to predict a bounded sequence of observations $y_1,\dots,y_n \in [0,B]$ ($B$ is unknown) using $E$ forecasting models called experts. For each time step $t \in \{ 1,\dots,n \}$, they provide $E$ forecasts $(\hat{y}_t^e)_{e=1}^E$ of the observation $y_t$. The aggregation $\hat{y}_t = \sum_{e=1}^E \hat{p}_{e,t} \hat{y}_t^e$ is then computed where the weights $(\hat{p}_{e,t})_{e=1}^E$ are updated online according to past performances of each expert. Forecast error is measured with a convex loss function $\ell_t(y_t, \cdot)$. The goal is to minimise the so-called regret $R_T = \frac{1}{T} \sum_{t=1}^T \ell_t(y_t, \hat{y}_t) - \frac{1}{T} \sum_{t=1}^T \ell_t(y_t, \hat{y}_t^\star)$ where $\hat{y}_t^\star$ is given by an oracle model which can use unavailable information to build a forecast difficult to beat. The regret is the difference between the error suffered by the aggregation and the error of the oracle. The latter can be the best-fixed convex combination of all the experts or the best-fixed expert (constant over time). We use the ML-Poly algorithm proposed by \cite{gaillard2014second} and implemented in the R package opera \citep{gaillard2016opera}. This algorithm tracks the best expert or the best convex combination of experts by giving more weight to an expert that will generate a low regret. This makes this algorithm particularly interesting as no parameter tuning is needed.

\subsection{Adaptive experts}\label{subsection:adaptiveexperts}

In this paragraph, we detail the forecasting methods we aggregate. We choose GAMs adapted by Kalman filtering as proposed by \cite{obst2021adaptive}. We consider in this section the forecasting of one time series.

\subsubsection{Generalized additive models}

We consider generalized additive models (GAM, \cite{wood2017generalized}) where the response variable $y_t$ is expressed as the following sum:
\begin{equation}
    y_t = \beta_0 + \sum_{d = 1}^D f_d(x_{t,d}) + \varepsilon_t,
\end{equation}
where $\beta_0$ is an intercept, $\varepsilon_t$ is the model error at time $t$, $(x_{t,d})_{d=1}^D$ are the $D$ explanatory variables available at time $t$ and $(f_d)_{d=1}^D$ are linear or nonlinear smooth functions called GAM effects. An effect $f_d$ is expressed as a projection

\begin{equation}
    f_d(x) = \sum_{k=1}^{m_d} \beta_{d,k} B_{d,k}(x),
\end{equation}
where $(B_{d,k})_{k=1}^{m_d}$ is a spline basis of dimension $m_d$ and $(\beta_{d,k})_{k=1}^{m_d}$ are the corresponding coefficients estimated by a ridge regression, where we minimize the following criterion:

\begin{equation}
    \sum_{t=1}^T \left( y_t - \sum_{d=1}^{D} f_d(x_{t,d}) \right)^2 + \sum_{d=1}^D \lambda_d \int \Vert f_d '' (x) \Vert^2 dx.
\end{equation}

The penalty term controls the second derivatives $f_d ''$ to force the effects to be smooth. We denote $C_1$ as the computational cost of GAM estimation.

\subsubsection{Adaptation by Kalman filter}

To adapt a GAM, we use a multiplicative correction of the following GAM effects vector $f(x_t) = (1, \overline{f}_1(x_{t,1}), \dots, \overline{f}_d(x_{t,d}))^\top$, where $\overline{f}_j$ is a normalized version of $f_j$ obtained by subtracting the mean on the train set and dividing by the standard deviation. 

The adaptation is obtained assuming that a state-space property is satisfied. Precisely, we estimate a vector $\theta_t$ called state under the assumption that
\begin{align}
    & y_t = \theta_t^\top f(x_t) + \varepsilon_t\,,\\
    & \theta_{t+1} = \theta_t + \eta_t\,,
\end{align}
where $(\varepsilon_t)$ and $(\eta_t)$ are Gaussian white noises of respective variance / covariance $\sigma^2$ and $Q$.

Starting from a Gaussian prior and assuming the variances $\sigma^2$ and $Q$ are known, the Kalman filter \citep{kalman1960new} achieves the estimation of the state $\theta_t$. This is a Bayesian method where at each step, the state posterior distribution is obtained as a Gaussian distribution $\theta_t\mid (x_s,y_s)_{s<t}\sim \mathcal{N}(\hat\theta_t,P_t)$.

The Kalman filter depends on the initial parameters $\hat{\theta}_1,P_1$ (the prior) and the variances $\sigma^2$ and $Q$. 
We choose these hyper-parameters by maximizing the likelihood on the training set. We use an iterative greedy procedure \cite[chapter 5]{de2022modeles}. We refer to this choice of hyper-parameters as the {\it dynamic} setting. Its computational cost is noted $C_2$ and can be large; it is detailed in the experimental study. Thanks to transfer learning, we avoid the costly estimation of these hyper-parameters for each time series. 

Another interesting setting, called {\it static}, is when $Q = 0$, $\sigma^2=1$, $P_1 = I$ and $\hat{\theta}_1 = 0$. In that case, the state equation becomes $\theta_{t+1}=\theta_t$ and the estimate $\hat{\theta}_t$ is equivalent to the ridge forecaster:
\begin{equation}
    \hat{\theta}_t = \argmin_{\theta \in \mathbb{R}^d} \left( \sum_{s=1}^{t-1} (y_s - \theta^\top f(x_s))^2 + \Vert \theta \Vert^2 \right).
\end{equation}

\subsubsection{Models considered}
\label{sec:models_considered}

As previously mentioned, the estimation of the GAM and of the Kalman variances is computationally costly when applied to 1,344 time series. We neglect the computational time due to Kalman updates, and we propose to reduce the number of GAMs and Kalman variances that are estimated.

We define hybrid experts $\mathcal{M}_{i,j,k}$, where the different parts of the forecaster are trained on different time series. GAM is trained on data set $i$, Kalman variances optimized (in the dynamic setting) on data set $j$, and resulting adapted model applied on data set $k$, where $i$, $j$, and $k$ range in $\{1,\dots,m_T\}$. We denote $\mathcal{M}_{i,\emptyset,k}$ a non-adapted GAM and $\mathcal{M}_{i,0,k}$ an adapted GAM in the static setting where no variances are optimized. 

There are three basic models without transfer learning where the models are optimized on the same substation's data we forecast: $\mathcal{M}_{k,k,k}$, $\mathcal{M}_{k,0,k}$, and $\mathcal{M}_{k,\emptyset,k}$. It is the scenario of traditional machine learning. We also consider three models involving transfer learning. The most simple transfer is GAM transfer: a source data set is used to train the GAM, and the resulting GAM is applied to a different target data set. It corresponds to the model $\mathcal{M}_{i,\emptyset,k},\ i \neq k$. An immediate improvement of the previous model is the adaptation of the transferred GAM with a Kalman filter optimized on the same data set used to train the GAM. It results in models $\mathcal{M}_{i,i,k}, \ i \neq k$. The adaptation step helps the GAM transfer and improves the basic GAM. Finally, we consider the Kalman filter transfer case: the data set we want to forecast is used to train a GAM adapted by a Kalman filter optimized on another data set. These are models $\mathcal{M}_{k,j,k},\ j \neq k$. 

\subsection{Aggregation models}\label{subsection:aggregationModels}

As previously said, the estimation of Kalman variances in the dynamic setting is computationally costly (e.g. training of $\mathcal{M}_{k,k,k}$ model), and we want to avoid individual training for all time series. Moreover, although the computational cost of GAM effects estimation is smaller, we possibly want to train only a few of them to reduce the cost and the number of model parameters. To do so, we build three aggregation methods based on the three previous models where transfer learning occurs, provided in Section \ref{sec:models_considered}. We obtain AGG GAM TL from the aggregation of $n_1~\mathcal{M}_{i,\emptyset,k}$ models, AGG GAM-Kalman TL from $n_2~ \mathcal{M}_{i,i,k}$ models, and AGG Kalman TL from $n_3~ \mathcal{M}_{k,j,k}$ models. The data sets used to train GAMs and Kalman filters in each aggregation method are randomly chosen among the 1,344 data sets.

Table \ref{table:description} details the three aggregation methods, and the two individual adapted GAMs. It specifies if a transfer is involved, the type of the model used, and the computational costs corresponding to GAM and Kalman variances estimation. More precisely, the computational costs correspond to GAM and Kalman variances estimations. Judging that GAM, Kalman filter, and aggregation applications are computationally cheap, we overlook them. For each aggregation method, parameter $n_i$ is the unique hyper-parameter of the method and corresponds to its number of sources $m_S$. It must be as small as possible compared to $m_T = 1344$ to force the frugality of the method. 

\begin{table}
    \centering
    \Large
    \resizebox{\columnwidth}{!}{\begin{tabular}{|>{\centering\arraybackslash}m{5cm}||>{\centering\arraybackslash}m{5cm}|>{\centering\arraybackslash}m{5cm}||>{\centering\arraybackslash}m{5cm}|>{\centering\arraybackslash}m{5cm}|>{\centering\arraybackslash}m{5cm}|}
        \hline
        Characteristics of adapted GAM & GAM + Kalman Static & GAM + Kalman Dynamic & AGG GAM TL & AGG GAM-Kalman TL & AGG Kalman TL \\
        \hline\hline
        Transfer of GAM & No & No & Yes & Yes & No \\
        Cost & $C_1 \times m_T$ & $C_1 \times m_T$  &$C_1 \times n_1 << C_1 \times m_T$ & $C_1 \times n_2 << C_1 \times m_T$ & $C_1 \times m_T$ \\ \hline
        Transfer of Kalman variances & - & No & - & Yes & Yes \\
        Cost & - & $C_2 \times m_T$ & - & $C_2 \times n_2 << C_2 \times m_T$ & $C_2 \times n_3 << C_2 \times m_T$ \\ 
        \hline
        Model type & $\mathcal{M}_{k,0,k}$ & $\mathcal{M}_{k,k,k}$ & $\mathcal{M}_{i,\emptyset,k}$ & $\mathcal{M}_{i,i,k}$ & $\mathcal{M}_{k,j,k}$ \\
        Total cost & $C_1 \times m_T$ & $(C_1 + C_2) \times m_T$ &  $C_1 \times n_1$ & $(C_1 + C_2) \times n_2$ & $C_1 \times m_T + C_2 \times n_3$ \\
        \hline
    \end{tabular}}
    \caption{Description of the individual adapted models and aggregation methods considered in the case study. $C_1$ and $C_2$ correspond to the computational costs of GAM and Kalman variances estimation, respectively. $m_T$ is the number of targets in the transfer learning context, and $n_i$ is the number of experts in the three aggregation methods. To give an order of magnitude, in our application, parameters $n_i$ are inferior to 10 while $m_T = 1,344$.}
    \label{table:description}
\end{table}

\section{Data and model presentation}\label{section:dataModelPresentation}

We first introduce French local electricity consumption and explanatory variables. We then detail the models in the practical point of view: definition of GAMs formula, training of Kalman filter and construction of the aggregation methods. To guarantee data confidentiality, substations' identities are not provided, and the electricity loads represented in the different figures are normalized by the average load.

\subsection{Presentation of the data}\label{subsection:dataPresentation}

\subsubsection{Electricity load data}

\begin{figure}
    \centering
    \includegraphics[scale=0.5]{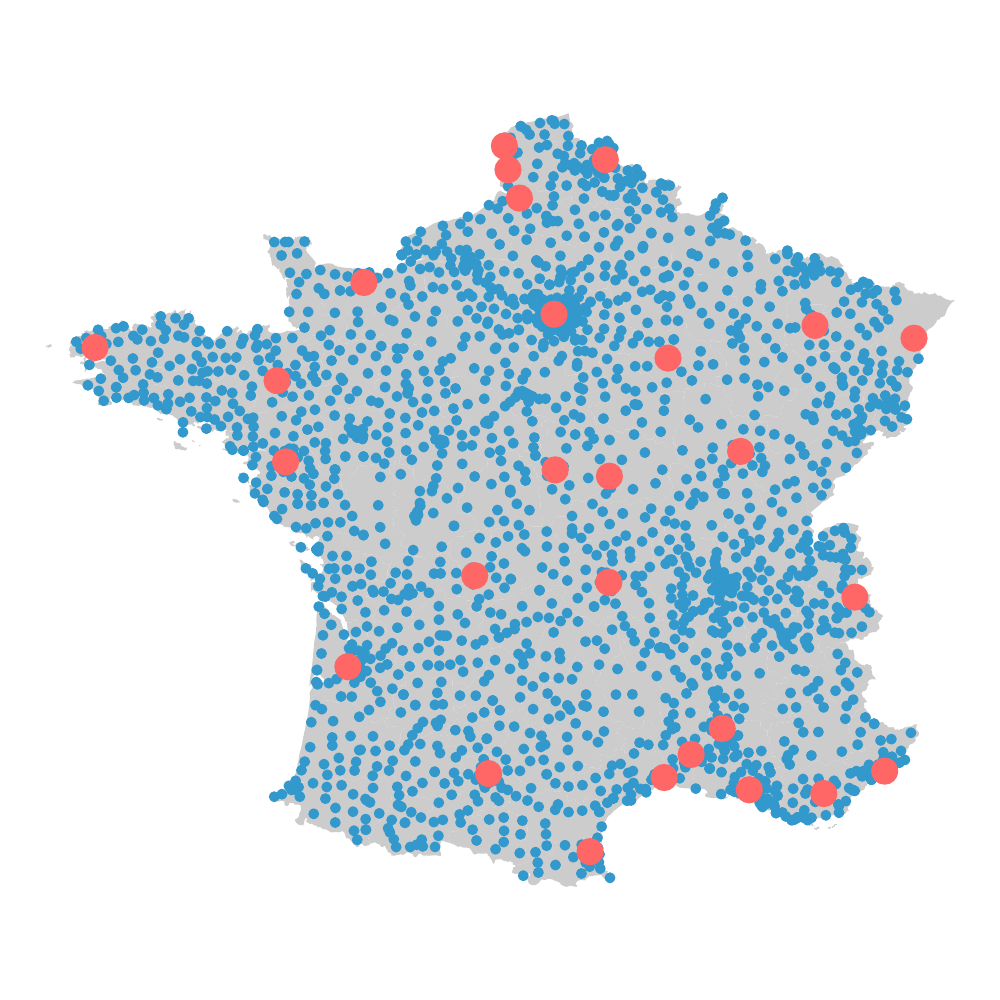}
    \caption{Location of the 1,344 substations (in blue) and 27 weather stations (in red).}
    \label{fig:franceStations}
\end{figure}

\begin{figure}
    \centering
    \includegraphics[width=\textwidth]{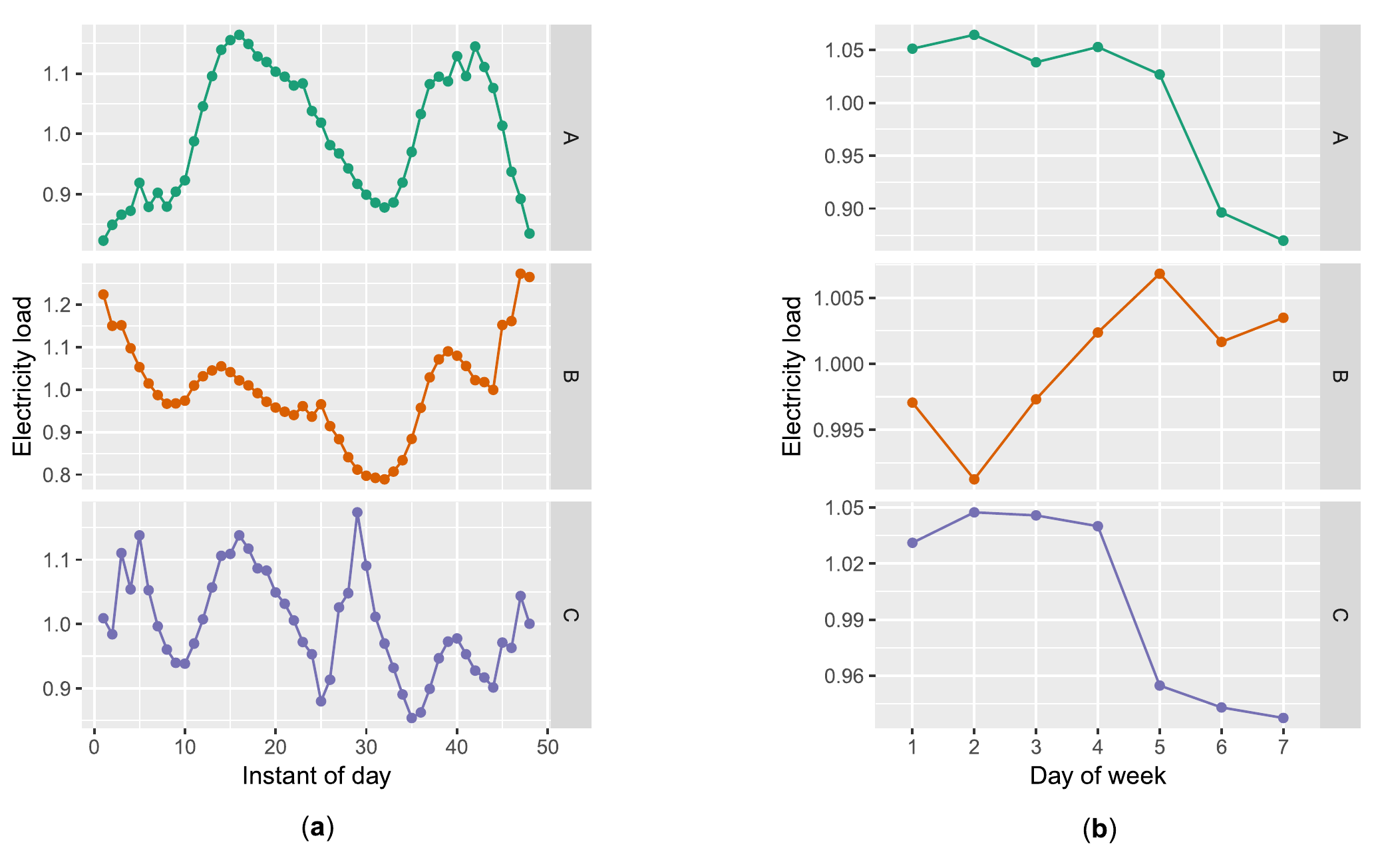}
    \caption{Electricity loads in March for three substations: (\textbf{a}) averaged daily load and (\textbf{b}) averaged weekly load. Substation A has the expected daily and weekly patterns: the load is ordinarily high during daylight, especially when people are at home, low during the night, and similar during weekdays, with a drop during weekends. On the other hand, substation B has opposed daily behavior at night, and load is increasing during the week and maximal during weekends. Substation C shows a high load in the mid-afternoon, and load is still important at night. Concerning its weekly load, it drops Friday in addition to the weekend.}
    \label{fig:dailyWeeklyLoad}
\end{figure}

The data are provided by Enedis, the operator in charge of the electric power distribution in France. The data are composed of 1,344 time series, each of which is the electricity consumption of one substation represented in blue in Fig. \ref{fig:franceStations}.
They cover metropolitan France and reflect thus its local electricity consumption. Forecasting each substation's consumption involves $m_T = 1,344$ forecasting tasks. The data are available from June 1\textsuperscript{st}, 2014, to December 31\textsuperscript{st}, 2021, with a 30-minute temporal resolution. We forecast electricity consumption for the next day with information up to the previous day; that’s why we fix an operational constraint on data availability with a lag of 48 hours. Although most of the 1,344 time series present classical temporal and meteorological patterns, there are some counter-intuitive and contrary variations. We refer to Fig. \ref{fig:dailyWeeklyLoad} for comparing one substation with basic behavior and two substations with unusual behaviors.

\subsubsection{Explanatory variables}

\begin{figure}
    \centering
    \includegraphics[width=\textwidth]{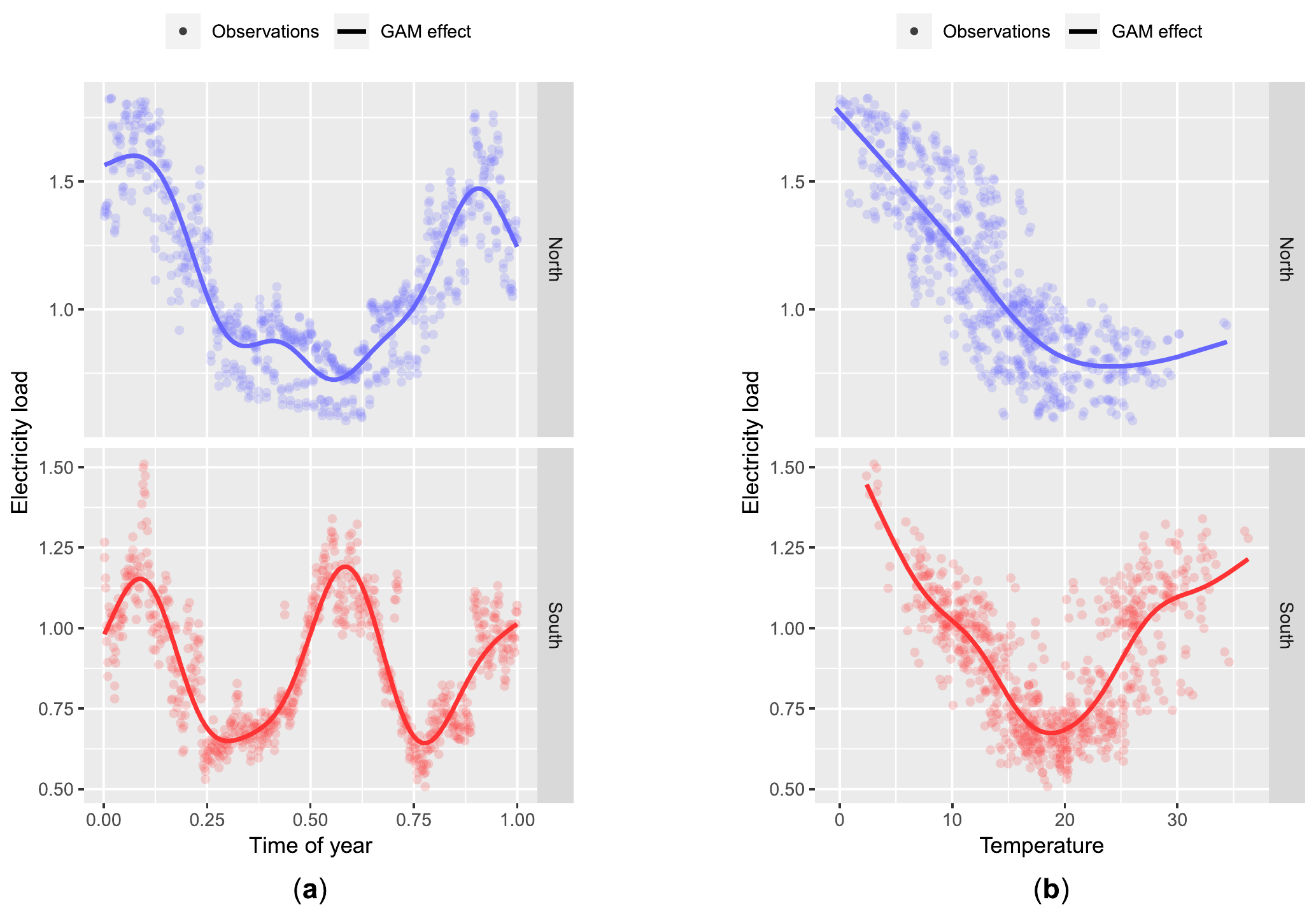}
    \caption{Dependence of 2015 electricity consumption at 6 p. m. to (\textbf{a}) time of year and (\textbf{b}) temperature. Blue values for a substation in the north of France and red values for one in the south of France.}
    \label{fig:dependanceTemperature}
\end{figure}

For explanatory variables, we choose meteorological and calendar variables, a common practice in electricity load forecasting. Weather variables are obtained from the French weather forecaster Météo-France. They are composed of temperature, cloudiness, and wind measurements from 27 weather stations. These stations are unequally spread over metropolitan France and are represented in red in Fig. \ref{fig:franceStations}. Weather data are available in the same range as electricity consumption with a 3 hours temporal resolution. We transform these data into 30 minutes of temporal resolution thanks to linear interpolation. Temperature is highly correlated to electricity load with a different impact for cold and hot regions, as shown in Fig. \ref{fig:dependanceTemperature}. The two patterns are due to the use of electrical heating and air-conditioning systems in France, which are highly electric consuming. On the other hand, calendar variables gather indicators of holidays by region, bank holidays, and working days, as well as the instant of the day, time of year, and days of the week.

\subsubsection{Time Segmentation}

\begin{figure}
    \centering
    \includegraphics[width = 0.6\textwidth, height = 0.4\textwidth]{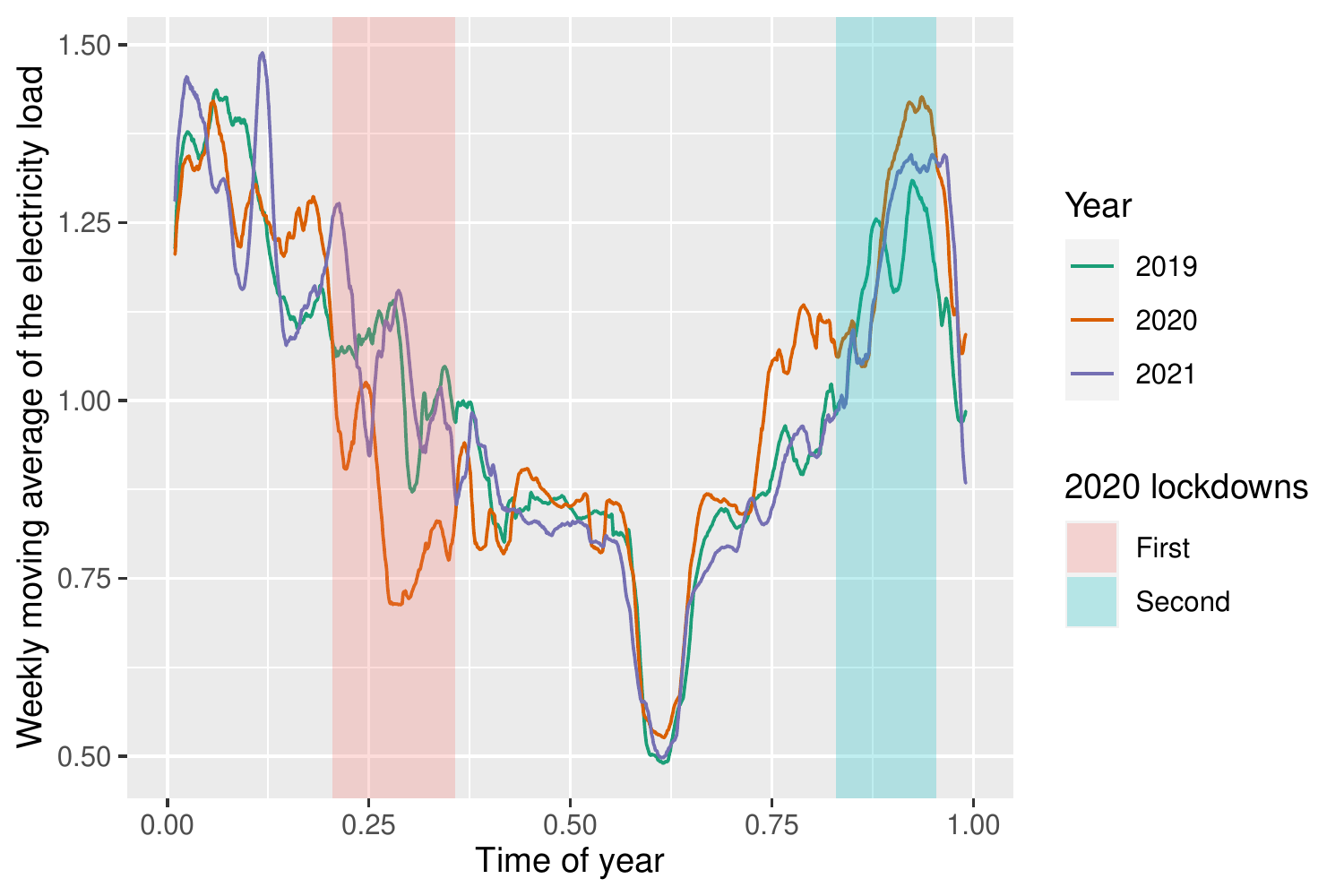}
    \caption{Weekly moving average of the electricity load of one substation in 2019, 2020, and 2021.}
    \label{fig:lockdownsLoad}
\end{figure}

The data have an availability range of 7.5 years. We train and validate the models on the data up to December 31\textsuperscript{st}, 2019, and we test afterward. The test set includes COVID-19 observations and three lockdown periods in France. We propose to consider the second and third lockdowns as normal periods because the electricity consumption strongly varies only during the first lockdown, see Fig. \ref{fig:lockdownsLoad}. We have thus chosen three validation periods: 2020 out of the first lockdown, the first lockdown (from March 16\textsuperscript{th}, 2020, to May 11\textsuperscript{th}, 2020), and 2021. 

\subsection{Generalized Additive Model formula}\label{subsection:GAMformula}

To build a GAM, we must determine its formula: which explanatory variables and spline bases to use. To do so, 2018 is used as a test period, and the different tests have been achieved with a forward-backward heuristic. We assume that a single GAM formula could be used for all target forecasting tasks. Indeed, although substations' behaviors are different, they can be explained by the same explanatory variables. Finally, we apply a GAM by instant of the day; that is, we deal with 48 GAMs for each forecasting task. We get the unique following GAM formula:

\begin{align*}
y_t = &\sum_{i=1}^5 \alpha_i 1_{DayType_t = i} + \sum_{i=1}^2 \beta_i 1_{BankHoliday_t = i} + \sum_{i=1}^2 f_i(ToY_t) 1_{WorkingDay=i} \\
& + f_3(Trend_t) + f_4(Temp_t) + f_5(Temp95_t) \\
&+ f_6(Temp99_t) + f_7(TempMin_t, TempMax_t) + \gamma_1 Load2D_t + \gamma_2 Load1W_t + \varepsilon_t,
\end{align*}

where at each time step $t$,

\begin{itemize}
    \item $y_t$ is the electricity load for the considered instant,
    \item $DayType_t$ is a categorical variable indicating the type of the day. There are five categories: Monday, Tuesday to Thursday, Friday, Saturday, and Sunday,
    \item $BankHoliday_t$ is a binary variable indicating whether the day $t$ is a bank holiday or a vacation day,
    \item $ToY_t$ is the time of year whose values grow linearly from 0 on the 1st of January midnight to 1 on the 31st of December 23h30,
    \item $WorkingDay_t$ is a binary variable indicating whether the day $t$ is a working day, i.e., not a weekend day or a bank holiday,
    \item $Trend_t$ is the number of the current observation,
    \item $Temp_t$ is the temperature of the closest weather station,
    \item $Temp95_t$ and $Temp99_t$ are exponentially smoothed $Temp_t$ variable of factor $\alpha = 0.95$ and $0.99$. E.g. for $\alpha = 0.95$ at a given time step $t$, $Temp95_t = \alpha Temp95_{t-1} + (1 - \alpha) Temp_t$,
    \item $TempMin_t$ and $TempMax_t$ are the minimal and maximal value of $Temp_t$ at the current day,
    \item $Load2D_t$ and $Load1W_t$ are loads of 2 days before and the load of the week before,
    \item $\varepsilon_t$ is Gaussian noise with 0 mean and constant variance. 
\end{itemize}

This model is a short-term model in terms of electricity past consumption availability. It is thus called ST GAM. We also consider a model in a mid-term context which is the same model without the $Load2D_t$ and $Load1W_t$ effects. It is called MT GAM.

We also examine the case where $\varepsilon_t$ is an auto-correlated error term. In that case, we use an ARIMA (autoregressive integrated moving average) model by selecting the best model with AIC criterion in the family of ARIMA(p,d,q). Correcting residuals of mid and short-term GAM with an ARIMA model achieves the same performance; therefore, we choose the more straightforward mid-term formula. It can be explained by the redundancy of correcting auto-correlated residuals with an ARIMA model and the linear lag terms in the short-term GAM. We call the following model MT GAM + ARIMA.

Thin plate spline basis with low dimensions represent all the effects except $f_1$ and $f_2$. Indeed, time of year has a cyclic impact (see Fig. \ref{fig:dependanceTemperature} (\textbf{a})); therefore, we use a cyclic cubic splines basis of dimension 20. GAM effects estimation is done using the Generalized Cross Validation criterion and takes a few tens of seconds in practice. Thus, it is computationally reasonable to train individual GAM for many forecasting tasks. Finally, we train the GAMs using R and the package mgcv \citep{wood2015package}.

\subsection{Kalman filtering}\label{subsection:kalman}

We examine in this section Kalman filtering adaptation of the previous short-term GAM. In the static setting, we can apply one individual adapted GAM to each forecasting task as no Kalman variances estimation is necessary. It is called GAM + Kalman Static. In this case, we are in a traditional machine learning context where the model is optimized on the same data set we forecast.

Execution time is essential in the dynamic setting. This is an unavoidable operational constraint, and we choose to calculate Kalman variances on only a subsample of the 1,344 substations. To select this subsample, we first use the short-term GAM predictions for 2018 to sort the 1,344 substations by forecast performance. The 44 substations corresponding to the 44 worst performances are put aside as we assume they won't provide interesting experts. We then randomly draw 6 substations in groups of 100 substations. We finally get a subsample of 78 substations representative of the forecasting difficulty. The parallel computing of the 78 corresponding sets of Kalman variances on a 36-core virtual machine on 2014-2018 data takes about 1.6 days. The computation of 1,344 individual Kalman variances would thus last about 28 days in a similar setting.

We calculate two sets of Kalman variances on the same substations subsample: one on 2014-2018 data and the other on 2014-2019 data. The first set is used to set the aggregation hyper-parameters $n_i$ with 2019 as the validation set, and the second set is used to forecast 2020-2021 as the test set. We thus have at most 78 GAMs adapted by their individual Kalman variances. In that case, we call the corresponding model GAM + Kalman Dynamic, and we show that the aggregation methods achieve equivalent performances.

Kalman filters estimation and application are achieved using R and the package viking \citep{viking2022package}. 

\subsection{Aggregation}\label{subsection:aggregation}

We refer to the Aggregation models section and Table \ref{table:description} for a remainder on the three aggregation methods. We focus here on their application.

\begin{figure}
    \centering
    \includegraphics[width=\textwidth]{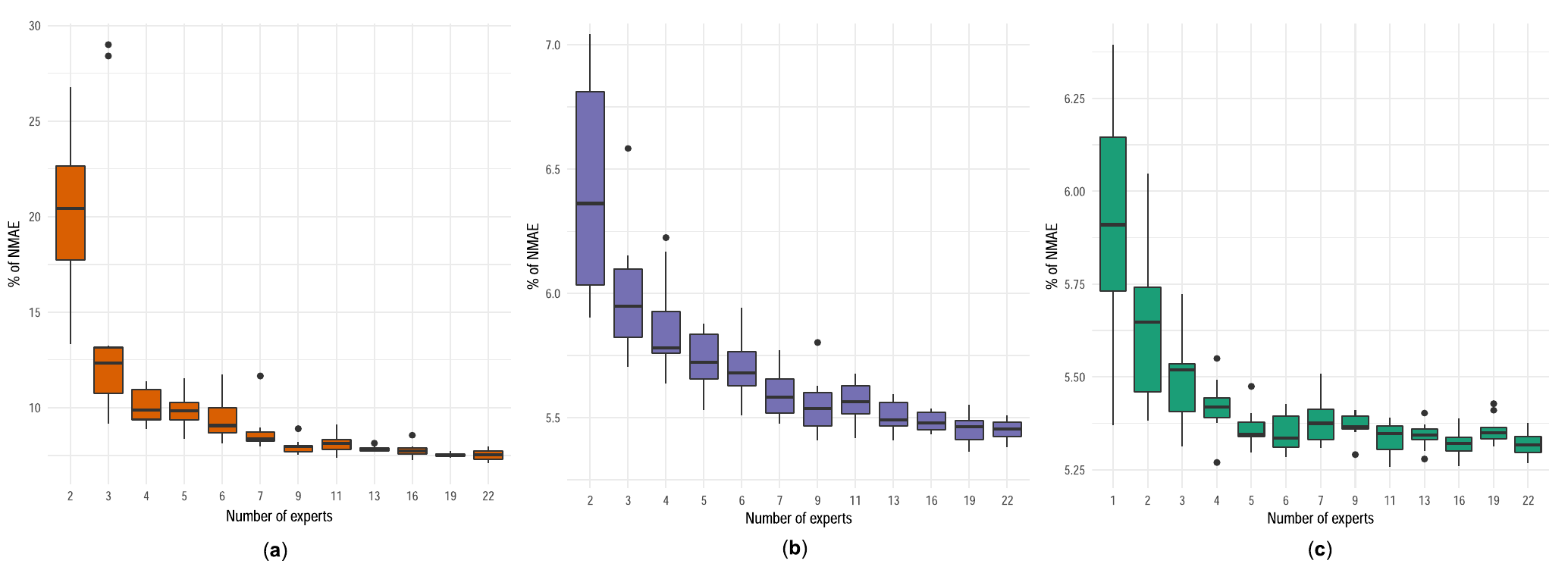}
    \caption{Grid search of the number of experts: (\textbf{a}) $n_1$ for AGG GAM TL, (\textbf{b}) $n_2$ for AGG GAM-Kalman TL, and (\textbf{c}) $n_3$ for AGG Kalman TL.}
    \label{fig:numberExperts}
\end{figure}

For each aggregation method, the unique hyper-parameter is $n_i$, the number of experts in the aggregation, which is also the number of sources in the transfer learning task. As said previously, we want $m_S$ to be the smallest possible compared to $m_T = 1,344$. We proceed with a grid search on 2019 data as the validation set. We perform 10 forecasts of 182 representative substations for different values of $n_i$ and compute the 10 corresponding medians. The 182 representative substations are obtained in the same way we obtain the 78 substations used to optimize Kalman variances, but here we pick 14 substations by groups of 100. We represent results in Fig. \ref{fig:numberExperts}. We can see that aggregation is gradually robust under the randomness of substations selection to compute GAM effects and Kalman variances. Moreover, there is an elbow phenomenon after $n_1 = n_2 = 9$ for AGG GAM TL and AGG GAM-Kalman TL, and after $n_3 = 6$ for AGG Kalman TL. These values of $n_i$ are very little compared to $m_T = 1,344$.

\section{Experiments}\label{section:experiments}

In this section, we first provide model analysis thanks to visualization plots and we then compare performances on substations data sets thanks to a comparative metric.

\subsection{Model dynamics}\label{subsection:dynamics}

\begin{figure}
    \centering
    \includegraphics[width=0.8\textwidth]{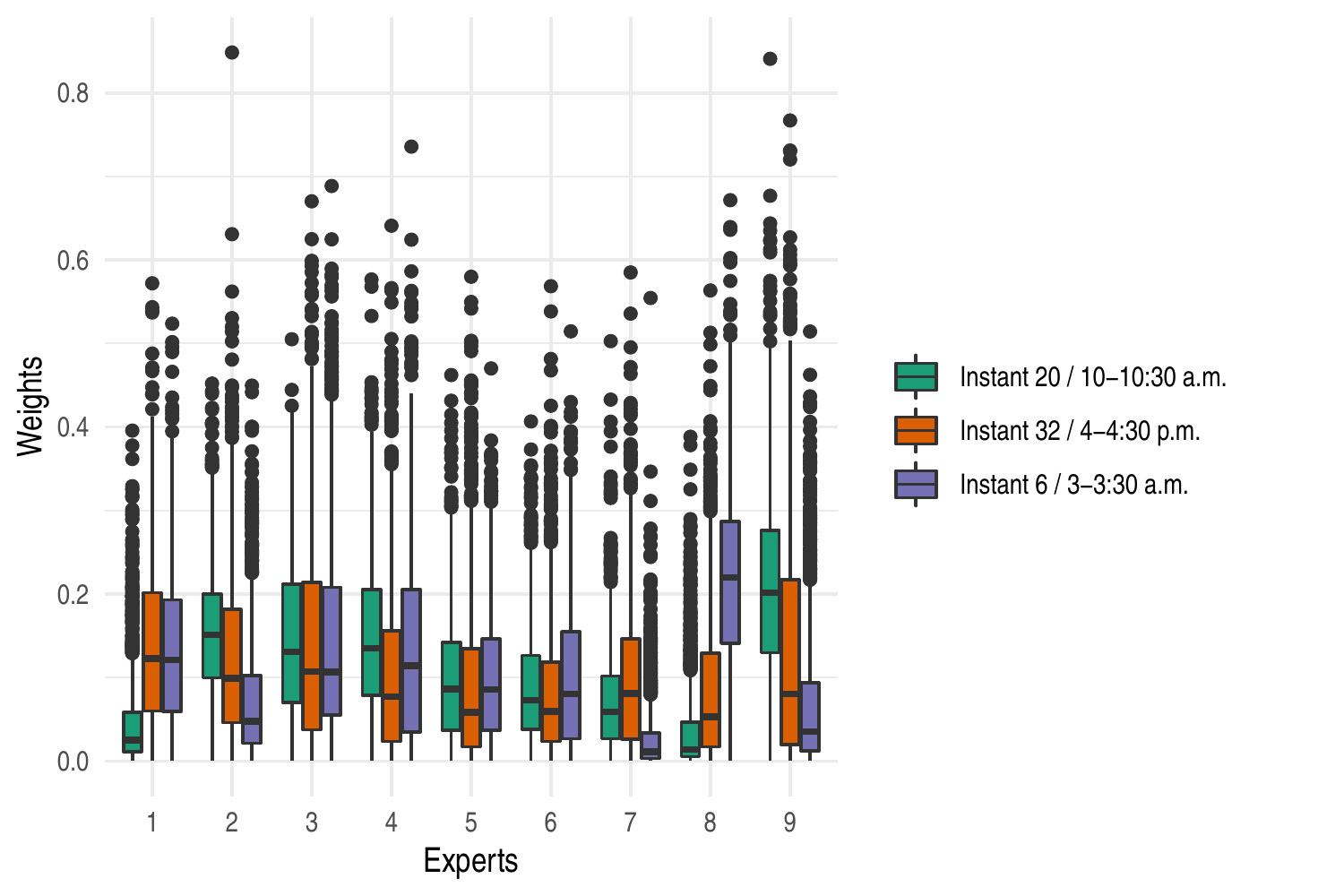}
    \caption{Boxplot representation of the weights of the 1,344 AGG GAM-Kalman TL aggregations for three instants of the day. 2021 target forecasting tasks.}
    \label{fig:expertsInstant}
\end{figure}

We first analyze the aggregation methods at the expert scale and then at the GAM effect scale. Concerning aggregations of experts, we want to know which expert is important in the mixture and when. To do so, we study the distribution of the weights of each expert. Indeed, the higher the weight, the more important the corresponding expert is in the forecast. One expert can be important in the forecast for specific observations and not for others. To give an example, we represent in Fig. \ref{fig:expertsInstant} boxplots of the weights of the 1,344 AGG GAM-Kalman TL aggregations for three interesting instants of the day where the consumption behaviors are very different. Experts 3, 4, 5, and 6 are essential in the same way according to the three instants of the day, with medians near the uniform weight of $1/9$. On the other hand, the other experts show variations according to the instant of the day. We can see that, for each instant, each expert is useless at least once (weight close to 0) and of great importance at least once (weight close to 0.5 or bigger).

\begin{figure}
    \centering
    \includegraphics[width=\textwidth]{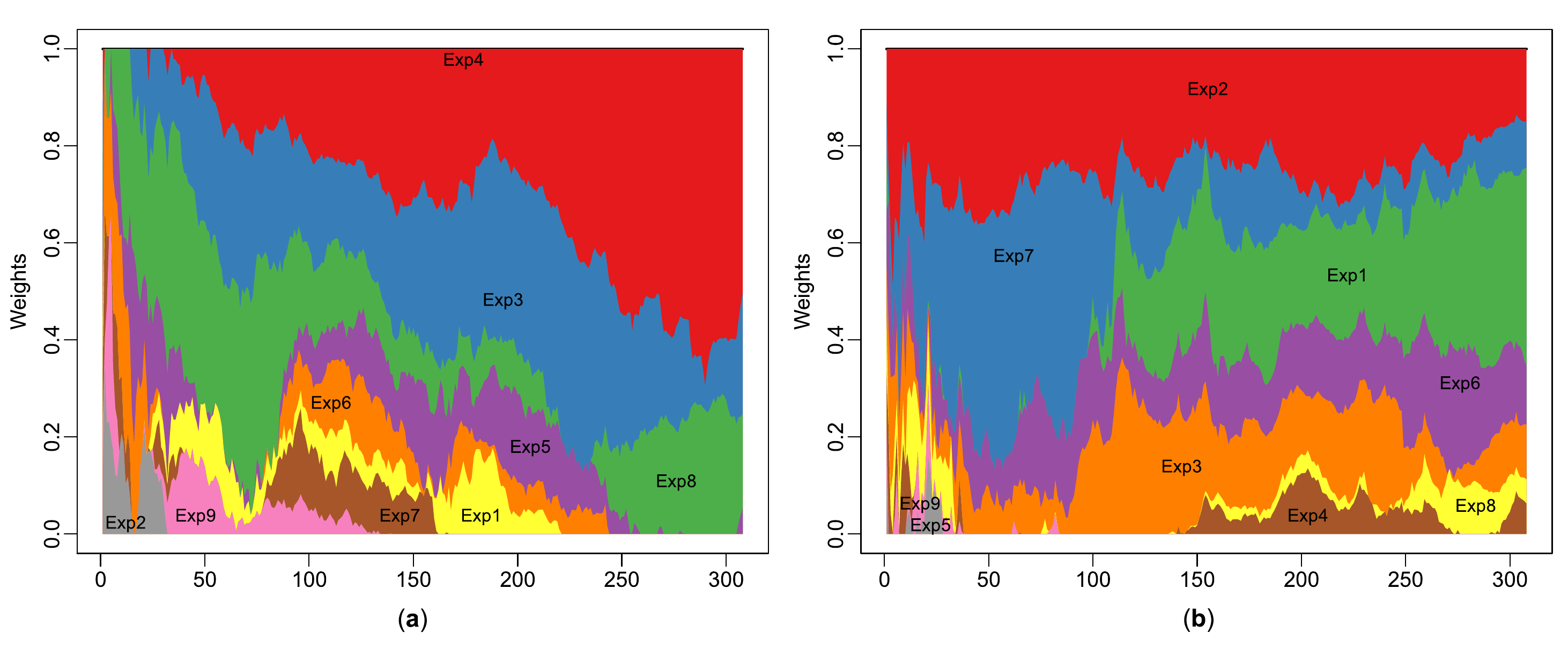}
    \caption{Weights associated with the experts of AGG GAM-Kalman TL aggregation at midday for (\textbf{a}) substation D and (\textbf{b}) substation E. Experts are denoted Exp$i$, and the forecasting period is 2020 out of the first lockdown.}
    \label{fig:Weights}
\end{figure}

We can focus on each forecast to see which experts are used and when in an aggregation. We display in Fig. \ref{fig:Weights} the evolution of weights from AGG GAM-Kalman TL aggregation. These figures represent the evolution of weights according to midday observations of 2 substations. We observe the same previous observations: some experts are called very little, while others are important. The new information these figures provide is the evolution of experts' importance during aggregation. For example, in Fig. \ref{fig:Weights} (\textbf{a}), Exp4's weight increases and becomes of paramount importance, with the weight being superior to $1/2$ for the last observations. In Fig. \ref{fig:Weights} (\textbf{b}), Exp7 vanishes rapidly while the impact of Exp1 increases. Experts 5 and 9 contribute very poorly to the first observations and are quickly absent in the aggregation.

\begin{figure}
    \centering
    \includegraphics[width=\textwidth]{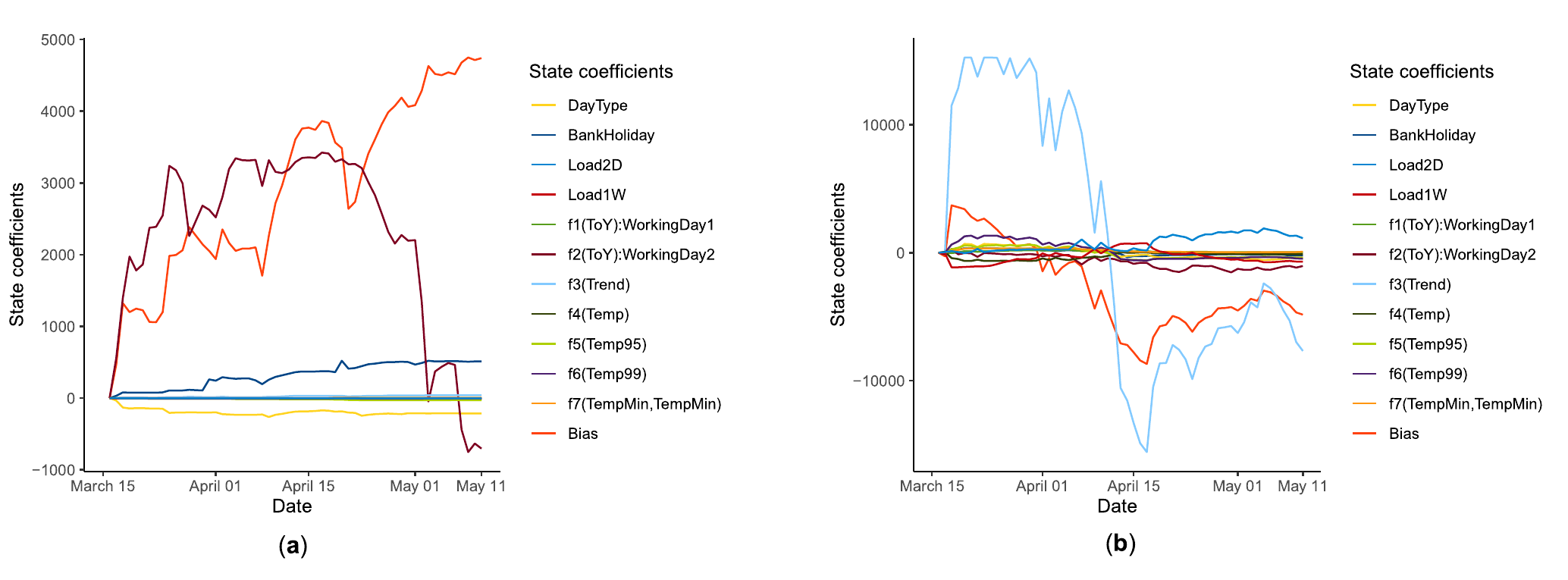}
    \caption{Evolution of (\textbf{a}) state coefficients of a Kalman TL model $(\hat{\theta}_{t,d} - \hat{\theta}_{0,d})_{d=1}^D$ and (\textbf{b}) hybrid state coefficients of a AGG Kalman TL aggregation $(\widetilde{\theta}_{t,d} - \widetilde{\theta}_{0,d})_{d=1}^D$, during the first lockdown and for the same target forecasting task.}
    \label{fig:theta}
\end{figure}

We also analyze the aggregation methods at the effect scale. To that end, we visualize the evolution of state coefficients from Kalman filtering; precisely, we plot the $D$ curves corresponding to the $D$ GAM effects. We provide an example in Fig. \ref{fig:theta} (\textbf{a}) with the evolution of state coefficients of an adapted GAM $\mathcal{M}_{k,k,k}$ during the first lockdown. That example is representative of what we observe for other state coefficients' evolution. Let's focus first on the coefficients associated with Bias and $f_2(ToY)$, which is the effect modeling the time of year during working days. Their coefficients are more significant compared to others and evolve rapidly. Concerning the bias coefficients, we think they evolve to balance the gap between the source load used to train the GAM and the target load, which can be of different scales. Then some state coefficients evolve significantly, like $BankHoliday$ or $DayType$, while others seem time-invariant.

If we select AGG Kalman TL, we can express the aggregation as a GAM adapted by hybrid state coefficients $(\widetilde{\theta}_{t,d})_{d=1}^D$:
\begin{equation}
    \hat{y}_t = \sum_{k=1}^E \hat{p}_{e,t} \hat{y}_{e,t} = \sum_{e=1}^E \hat{p}_{e,t} \sum_{d=1}^D \hat{\theta}_{t,d,e} f_d (x_{t,d}) = \sum_{d=1}^D \left( \sum_{k=1}^E \hat{p}_{e,t} \hat{\theta}_{t,d,e} \right) f_d(x_{t,d}) = \sum_{d=1}^D \widetilde{\theta}_{t,d} f_d(x_{t,d}),
\end{equation}
where at time $t$ we denote $\widetilde{\theta}_{t,d} = \sum_{e=1}^E \hat{p}_{e,t} \hat{\theta}_{t,d,e}$ the hybrid state coefficients, $E$ the number of experts, $D$ the number of GAM effects, $\hat{p}_{e,t}$ the weight associated with expert $e$, and $\hat{\theta}_{t,d,e}$ the adaptation coefficient associated with the GAM effect $f_d$ of expert $e$. This new expression allows for gathering information on both adaptation and aggregation and improving the interpretation of the whole model. We represent in Fig. \ref{fig:theta} (\textbf{b}) the evolution of these hybrid state coefficients during the first lockdown for the same target forecasting task as in Fig. \ref{fig:theta} (\textbf{a}). We can therefore compare the Kalman coefficients $(\hat{\theta}_{t,d,e})_{d=1}^D$ of an individual model $\mathcal{M}_{k,k,k}$ and the hybrid state coefficients $(\widetilde{\theta}_{t,d})_{d=1}^D$ calculated from $n_3 ~ \mathcal{M}_{k,j,k}$ models coefficients. We see that both types of coefficients differ in amplitude and dynamic. Bias coefficients are still important compared to the others and evolve dynamically, but it's not the case for $f_2$ coefficients. Moreover, most of the hybrid state coefficients evolve contrary to the previous ones. Finally, we can make the same observations of this example for the other substations.

If we consider now any aggregation method, we can express the whole mixture as a GAM using the linearity of GAM effects, GAMs, the adaptation procedure, and the aggregation of experts. The new coefficients of the spline basis are composed of the basic coefficients of the spline basis, the adaptation coefficients, and the aggregation weights. 

\subsection{Numerical results}\label{subsection:numericalResults}

To compute numerical performances of the models, we use the normalized mean absolute error (NMAE) defined as 
\begin{equation}
    NMAE(y,\hat{y})=\frac{\sum_{i=1}^n |y_i - \hat{y}_i|}{\sum_{i=1}^n |y_i|}.
\end{equation}
It is the mean absolute error between the ground truth $y$ and the forecast $\hat{y}$ normalized by the absolute mean of $y$. This metric allows us to compare forecasts of different scales, which is necessary in our case study. We choose the NMAE rather than the mean absolute percentage error because our time series are sometimes very close to 0. More precisely, we use the percentage of NMAE by multiplying it by 100.

Table \ref{table:perfs1,344} provides the numerical performances of the 1,344 forecasting tasks. For each model, the first, second, and third quartiles of the 1,344 performances are provided during the three validation periods. Kalman variances and GAMs have been trained on 2014-2019 for each test period. We display with the same setting in Table \ref{table:perfs78} the numerical performances of the forecasting tasks corresponding to the substations subsample chosen to optimize the set of Kalman variances.

We start by comparing the performances of the individual and aggregation models displayed in Table \ref{table:perfs1,344}. Each time we cite improvements in the performances, we refer to the three respective validation periods. We can see that the information on the past electricity load added in the GAM formula is of great importance, with an improvement between ST GAM and MT GAM median performances of 1.84\%, 2.83\%, and 3.13\%. There is also an improvement when MT GAM residuals are modeled with an ARIMA model: there is an improvement between MT GAM and MT GAM + ARIMA median performances of 0.77\%, 1.46\%, and 1.70\%. GAM + Kalman Static provides slightly weaker performances than MT GAM + ARIMA, and the latter is thus the individual benchmark to beat. 

\begin{figure}
    \centering
    \includegraphics[width=0.6\textwidth]{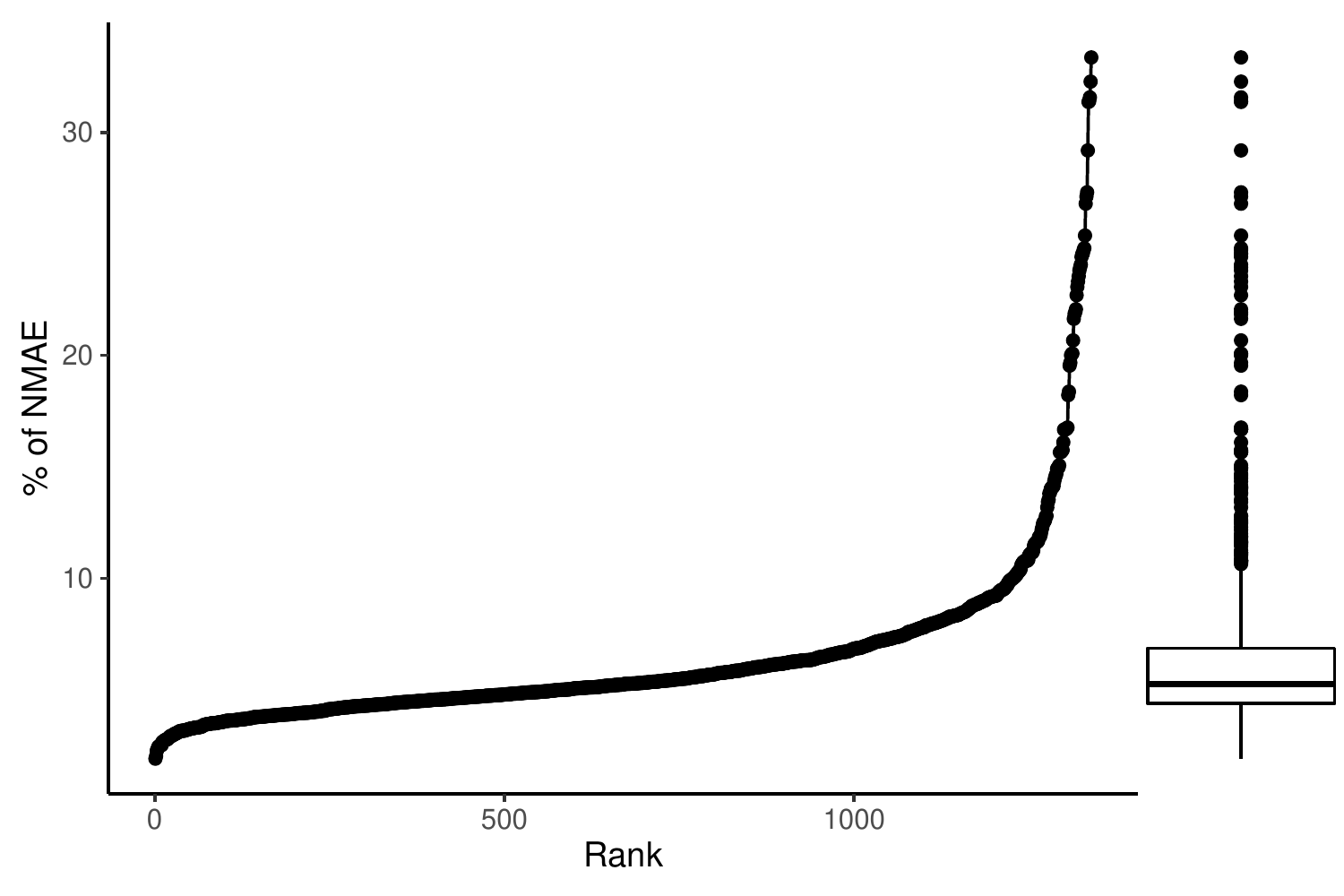}
    \caption{Sorted performances of AGG Kalman TL for the 1,344 forecasting tasks in 2021.}
    \label{fig:curveResults}
\end{figure}

Concerning the aggregation methods, AGG GAM TL shows the worst performance. However, this model is still interesting as its performances approach the ST GAM performances while its computational cost is much lower. On the other hand, the two other aggregation methods show excellent and close performances. AGG Kalman TL is better than AGG GAM-Kalman TL outside of the first lockdown (improvement in the median of 0.03\% and 0.07\%), while it is the opposite for the first lockdown (decrease of 0.16\%). The gap between these two last models is not significant. As AGG GAM-Kalman TL is less complex than AGG Kalman TL, we consider it the best compromise between forecasting performance and computational cost. Finally, compared to MT GAM + ARIMA, its median performances improve by 0.62\%, 1.54\%, and 0.84\%. We represent its 1,344 NMAE scores in Fig. \ref{fig:curveResults}. 

As explained in Section \ref{subsection:aggregationPrinciple}, aggregation algorithms aim at beating two oracles: the best-fixed expert and the best-fixed convex combination of experts. We call them Expert Oracle and Convex Oracle, respectively, and we compute their performances for AGG GAM-Kalman TL. The aggregation competes with the best-fixed expert and approaches the best-fixed combination of experts. We observe the same phenomenon for the two other aggregation methods.

Concerning the test periods, we can see that the 2020 forecasts worsen during the first French lockdown with a downgrade of 2.18\% between the two MT GAM + ARIMA median performances. AGG GAM-Kalman TL and AGG Kalman TL present lower downgrades: 1.26\% and 1.45\% in the median, respectively; that is, the online nature of Kalman filtering and aggregation of experts allow adaptation to the extreme change in electricity consumption. On the other hand, ST GAM and MT GAM performances worsen in 2021 as these models are not updated while the consumption behavior evolves. The downgrades are significant: 1.37\% and 2.36\% in the median, respectively. The performance gaps between 2020 out of the first lockdown and 2021 are tighter for adaptative models, especially for the two best aggregation methods: 0.22\% and 0.18\% in the median, respectively. Once again, it shows that the aggregation methods adapt well to the data distribution evolution.

\begin{table}
    \centering
    \resizebox{\columnwidth}{!}{\begin{tabular}{|c|c c c|c c c|c c c|}
        \hline
        Method & \multicolumn{3}{|c|}{2020 out of the first lockdown} & \multicolumn{3}{|c|}{First lockdown} & \multicolumn{3}{|c|}{2021} \\
        \hline\hline
        ST GAM &  5.35 \% & 6.50 \% & ~8.69 \% &  7.31 \% & ~9.37 \% & 13.30 \% &  6.29 \% & ~7.87 \% & 11.00 \% \\
        MT GAM &  6.67 \% & 8.34 \% & 11.50 \% &  9.44 \% & 12.50 \% & 18.70 \% &  ~8.17 \% & 10.70 \% & 16.30 \% \\
        MT GAM + ARIMA &  4.74 \% & 5.73 \% & ~7.53 \% &  6.28 \% & ~7.91 \% & 10.70 \% &  5.13 \% & ~6.17 \% & ~8.37 \% \\
        \hline
        GAM + Kalman Static & 5.00 \% & 5.98 \% & ~7.90 \% & 6.96 \% & ~8.84 \% & 12.20 \% & 5.26 \% & ~6.25 \% & ~8.32 \% \\
        \hline
        AGG GAM TL & 6.21 \% & 7.67 \% & 10.90 \% & 8.02 \% & 10.10 \% & 14.40 \% & 6.23 \% & ~7.58 \% & 10.80 \% \\
        AGG GAM-Kalman TL &  4.29 \% & 5.11 \% & ~6.68 \% & 5.26 \% & ~6.37 \% & ~8.20 \% & 4.44 \% & ~5.33 \% & ~6.98 \% \\
        AGG Kalman TL & 4.25 \% & 5.08 \% & ~6.70 \% & 5.43 \% & ~6.53 \% & ~8.56 \% & 4.39 \% & ~5.26 \% & ~6.87 \% \\
        \hline
        Expert Oracle & 4.40 \% & 5.24 \% & ~6.78 \% & 5.01 \% & ~5.98 \% & ~7.67 \% & 4.58 \% & ~5.47 \% & ~7.07 \% \\
        Convex Oracle & 4.11 \% & 4.90 \% & ~6.32 \% & 4.52 \% & ~5.41 \% & ~6.97 \% & 4.26 \% & ~5.12 \% & ~6.67 \% \\
        \hline
    \end{tabular}}
    \caption{Numerical performances of the 1,344 forecasting tasks in normalized mean absolute error (NMAE) (\%). Format 1\textsuperscript{st} quartile ~~ Median ~~ 3rd quartile.}
    \label{table:perfs1,344}
\end{table}

\begin{table}
    \centering
    \resizebox{\columnwidth}{!}{\begin{tabular}{|c|c c c|c c c|c c c|}
        \hline
        Method & \multicolumn{3}{|c|}{2020 out of the first lockdown} & \multicolumn{3}{|c|}{First lockdown} & \multicolumn{3}{|c|}{2021} \\
        \hline\hline
        GAM + Kalman Dynamic &  4.45 \% & 5.36 \% & ~7.02 \% & ~5.75 \% & ~7.50 \% & 10.10 \% &  4.55 \% & ~5.22 \% & ~7.79 \% \\
        \hline
        AGG GAM TL &  6.25 \% & 7.79 \% & 11.30 \% & ~8.65 \% & 11.10 \% & 14.90 \% &  6.05 \% & 7.63 \% & 13.10 \% \\
        AGG GAM-Kalman TL &  4.23 \% & 5.36 \% & ~7.11 \% & ~5.64 \% & ~6.91 \% & ~8.86 \% &  4.27 \% & ~5.41 \% & ~7.73 \% \\
        AGG Kalman TL & 4.39 \% & 5.48 \% & ~7.15 \% & ~6.03 \% & ~7.67 \% & ~10.10 \% &  4.48 \% & ~5.42 \% & ~7.89 \% \\
        \hline
    \end{tabular}}
    \caption{Numerical performances of 69 representative forecasting tasks in normalized mean absolute error (NMAE) (\%). Format: 1st quartile ~~ Median ~~ 3rd quartile.}
    \label{table:perfs78}
\end{table}

Finally, we provide in Table \ref{table:perfs78} performances of 69 forecasting tasks. Precisely, they correspond to the 78 substations we used to optimize the set of Kalman variances out of the 9 substations data sets selected to train the experts involved in the three aggregation methods. That is we avoid the lucky case where Kalman variances are optimized on a data set and used to forecast this data set. We can thus compare the aggregation methods with GAM + Kalman Dynamic in the most common case. As previously said, GAM + Kalman Dynamic achieves great performance forecasting the national electricity load and is thus a model we want to compete with. We can see that it is the case for AGG GAM-Kalman TL and AGG Kalman TL; therefore, these 2 methods are accurate comparing to the state-of-the-art model.

\section{Conclusion}\label{section:conclusion}

In this paper, we propose a frugal method to forecast multiple electricity loads. We avoid the estimation of an individual model per time series and we obtain a scalable methodology with limited resources based on aggregation of experts. The chosen experts are GAMs adapted by Kalman filtering, which are state-of-the-art models in electricity consumption forecasting. The aggregation methods allow the transfer of GAMs and Kalman filters separately or simultaneously. We show that they provide good forecasts compared to classical individual models, especially during the first French lockdown. Moreover, they compete with individual state-of-the-art GAM adapted by Kalman filtering. On the other hand, they don't need human intervention nor expertise and are thus simple to use. The method is frugal in terms of parameter estimation, and the computational cost has been discussed extensively. Finally, it benefits from the interpretability of GAM and the aggregation of experts. 

We see various ways to extend our work. First, we think we can improve the random selection of the time series used to train GAMs and Kalman variances. We may characterize clusters of substations according to some characteristics (geographic, weather, type of consumers) and select one representative per cluster. The second way of improvement is the inclusion of new explanatory variables at the same local scale. Geo-tracking and communication data reflect human behavior and are therefore helpful for electric consumption forecasting \citep{gaucher2021hierarchical}. Finally, the computational complexity of the method introduced essentially depends on the estimation of a fixed number of models; therefore, we could apply the method to a larger data set, for instance the electrical consumption at a finer granularity.

\bibliography{references}

\end{document}